\documentclass{article} 
\usepackage{iclr2026_conference,times}

\usepackage{amsmath,amsfonts,bm}

\def\eqref#1{equation~\ref{#1}}

\def\1{\bm{1}}

\DeclareMathAlphabet{\mathsfit}{\encodingdefault}{\sfdefault}{m}{sl}
\SetMathAlphabet{\mathsfit}{bold}{\encodingdefault}{\sfdefault}{bx}{n}

\usepackage{hyperref}
\usepackage{url}
\usepackage{amsmath, amssymb}
\usepackage{graphicx}
\usepackage{booktabs}
\usepackage{enumitem}
\usepackage{algorithm}
\usepackage{algpseudocode}
\usepackage{tcolorbox}
\tcbuselibrary{listings,breakable,skins}

\definecolor{darkblue}{RGB}{0, 70, 140}

\usepackage{xstring}
\definecolor{cerrgreen}{RGB}{0,100,0}     
\definecolor{cerrred}{RGB}{160,0,0}      

\newcommand{\cerr}[1]{%
  \IfSubStr{#1}{e}{%
    \StrBefore{#1}{e}[\cerrmant]%
    \StrBehind{#1}{e}[\cerrexp]%
    \ifnum\cerrexp<-4\relax \textcolor{cerrgreen}{$\cerrmant\times10^{\cerrexp}$}%
    \else\ifnum\cerrexp<-1\relax $\cerrmant\times10^{\cerrexp}$%
    \else \textcolor{cerrred}{$\cerrmant\times10^{\cerrexp}$}%
    \fi\fi
  }{%
    \textcolor{cerrred}{$#1$}%
  }%
}
\newcommand{\cerrb}[1]{%
  \IfSubStr{#1}{e}{%
    \StrBefore{#1}{e}[\cerrmant]%
    \StrBehind{#1}{e}[\cerrexp]%
    \ifnum\cerrexp<-4\relax \textcolor{cerrgreen}{$\mathbf{\cerrmant\times10^{\cerrexp}}$}%
    \else\ifnum\cerrexp<-1\relax $\mathbf{\cerrmant\times10^{\cerrexp}}$%
    \else \textcolor{cerrred}{$\mathbf{\cerrmant\times10^{\cerrexp}}$}%
    \fi\fi
  }{%
    \textcolor{cerrred}{$\mathbf{#1}$}%
  }%
}

\title{AutoNumerics: An Autonomous, PDE-Agnostic Multi-Agent Pipeline for Scientific Computing}

\author{%
Jianda Du$^{1}$ \quad Youran Sun$^{1}$ \quad Haizhao Yang$^{1,2,*}$ \\[0.6ex]
$^{1}$Department of Mathematics, University of Maryland, College Park, MD, USA \\
$^{2}$Department of Computer Science, University of Maryland, College Park, MD, USA \\[0.4ex]
\texttt{jdu37576@umd.edu} \quad \texttt{sun1245@umd.edu} \quad \texttt{hzyang@umd.edu}
}

\iclrfinalcopy 
\begin{document}

\maketitle
\lhead{}
\begingroup
\renewcommand\thefootnote{*}
\footnotetext{Corresponding author.}
\endgroup

\begin{abstract}
PDEs are central to scientific and engineering modeling, yet designing accurate numerical solvers typically requires substantial mathematical expertise and manual tuning.
Recent neural network-based approaches improve flexibility but often demand high computational cost and suffer from limited interpretability.
We introduce \texttt{AutoNumerics}, a multi-agent framework that autonomously designs, implements, debugs, and verifies numerical solvers for general PDEs directly from natural language descriptions.
Unlike black-box neural solvers, our framework generates transparent solvers grounded in classical numerical analysis.
We introduce a coarse-to-fine execution strategy and a residual-based self-verification mechanism.
Experiments on 24 canonical and real-world PDE problems demonstrate that \texttt{AutoNumerics} achieves competitive or superior accuracy compared to existing neural and LLM-based baselines, and correctly selects numerical schemes based on PDE structural properties, suggesting its viability as an accessible paradigm for automated PDE solving.
\end{abstract}

\section{Introduction}
Partial differential equations (PDEs) form the mathematical foundation of modern physics, engineering, and many areas of scientific computing.
Accurately solving PDEs is therefore a central task in computational research.
Traditionally, constructing a reliable numerical solver for a new PDE requires substantial expertise in numerical analysis, including the selection of appropriate discretization schemes (e.g., finite difference, finite element, or spectral methods) and verification of stability and convergence conditions such as the Courant--Friedrichs--Lewy (CFL) constraint~\citep{leveque2007finite}.
These classical approaches provide strong mathematical guarantees and interpretability, but their expert-driven design can limit accessibility and slow solver development for newly arising PDE models.

Neural network-based approaches such as physics-informed neural networks (PINNs)~\citep{raissi2019physics} and operator-learning frameworks~\citep{lu2019deeponet,li2020fourier} reduce reliance on handcrafted discretizations but introduce new concerns around computational cost and interpretability.
Large language models (LLMs) have recently demonstrated strong capabilities in scientific code generation~\citep{zhang2024comprehensive}, and existing LLM-assisted PDE efforts include neural solver design~\citep{he2025lang,jiang2025agenticsciml}, tool-oriented systems that invoke libraries such as FEniCS~\citep{liu2025pde,wu2025automated}, and code-generation paradigms~\citep{li2025codepde}.
However, these approaches either produce black-box networks, are constrained by fixed library APIs, or lack mechanisms for autonomous debugging and correctness verification.
We propose that LLMs can serve as \emph{numerical architects} that directly generate transparent solver code from first principles, preserving interpretability while automating solver construction.

Translating this vision into a reliable system poses several technical challenges.
First, LLM-generated code often contains syntax errors or logical flaws, and debugging these errors on high-resolution grids is both time-consuming and computationally wasteful.
Second, verifying solver correctness becomes difficult for PDEs lacking analytical solutions.
Third, large-scale temporal simulations may lead to memory exhaustion.
We address these challenges with three corresponding solutions.
A coarse-to-fine execution strategy first debugs logic errors on low-resolution grids before running on high-resolution grids.
A residual-based self-verification mechanism evaluates solver quality for problems without analytical solutions by computing PDE residual norms.
A history decimation mechanism enables large-scale temporal simulations through sparse storage of intermediate states.

Building on these design principles, we propose \texttt{AutoNumerics}, a multi-agent autonomous framework.
The system receives natural language problem descriptions, proposes multiple candidate numerical strategies through a planning agent, implements executable solvers, and systematically evaluates their correctness and performance.
We evaluate the framework on 24 representative PDE problems spanning canonical benchmarks and real-world applications.
Results demonstrate consistent numerical scheme selection, stable solver synthesis, and reliable accuracy across diverse PDE classes.

\paragraph{Position relative to prior work.}
Existing LLM-assisted PDE efforts include neural solver design~\citep{he2025lang,jiang2025agenticsciml}, tool-oriented systems that invoke libraries such as FEniCS~\citep{liu2025pde,wu2025automated}, and code-generation paradigms~\citep{li2025codepde}.
\texttt{AutoNumerics} differs from all three.
It generates interpretable classical numerical schemes (not black-box networks), automatically detects and filters ill-designed or non-expert numerical plan configurations, derives discretizations from first principles (not fixed library APIs), and includes a coarse-to-fine execution strategy with residual-based self-verification for autonomous correctness assessment.
A detailed review of related work is provided in Appendix~\ref{sec:related-work}.

\paragraph{Contributions.}
The primary contributions of this work are:
\begin{itemize}[nosep,leftmargin=1.5em]
    \item A multi-agent framework (\texttt{AutoNumerics}) that autonomously constructs transparent numerical PDE solvers from natural language descriptions.
    \item A reasoning module that detects ill-designed or non-expert PDE specifications and proactively filters or revises numerical plans that may lead to instability or invalid solutions.
    \item A coarse-to-fine execution strategy that decouples logic debugging from stability validation.
    \item A residual-based self-verification mechanism for solver evaluation without analytical solutions.
    \item A benchmark suite of 200 PDEs and systematic evaluation on 24 representative problems, with comparisons to neural network baselines and CodePDE.
\end{itemize}

\section{Method}

\subsection{Problem Formulation and Plan Generation}

\begin{figure}[t]
\centering
\includegraphics[width=\textwidth]{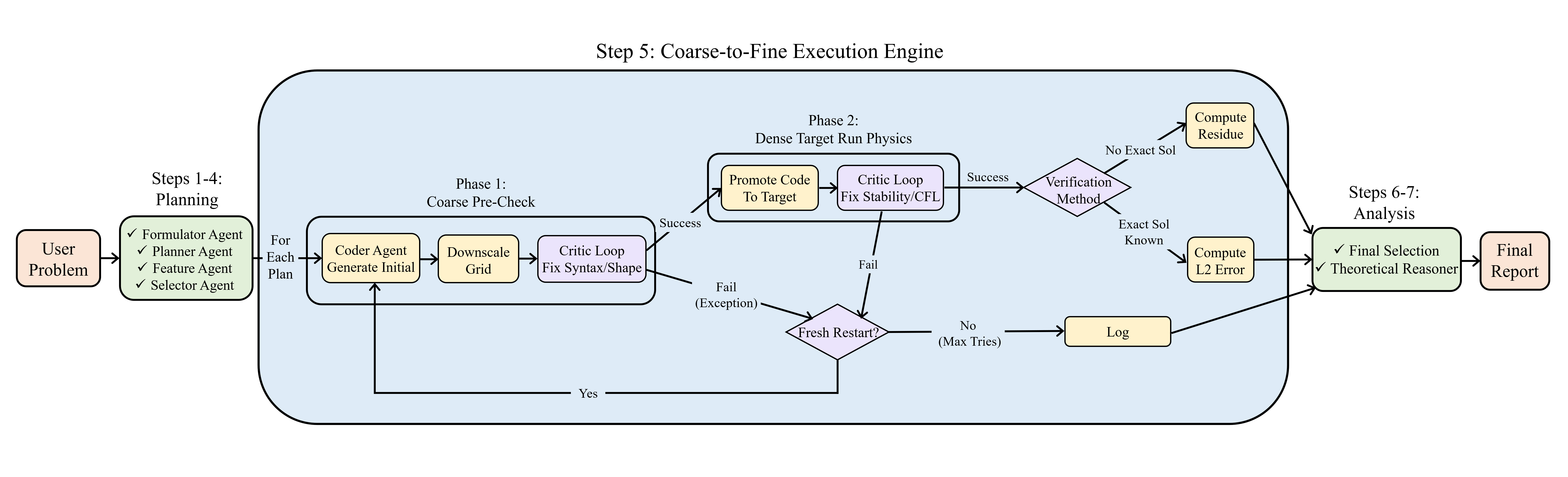}
\caption{The AutoNumerics pipeline.
Steps 1--4 handle problem formulation and plan selection.
Step 5 implements the coarse-to-fine execution strategy with Fresh Restart logic.
Steps 6--7 perform verification and theoretical analysis.}
\label{fig:flowchart}
\end{figure}

AutoNumerics consists of multiple specialized LLM agents coordinated by a central dispatcher.
The system takes a natural language PDE problem description as input and produces executable numerical solver code with accuracy metrics as output.
The overall architecture is illustrated in Figure~\ref{fig:flowchart}.

The pipeline begins with the Formulator Agent, which converts the natural language description into a structured specification containing governing equations, boundary and initial conditions, and physical parameters.
The Planner Agent then proposes multiple candidate schemes covering different discretization methods (e.g., finite difference, spectral, finite volume) and time-stepping strategies (explicit, implicit), while avoiding configurations
that violate basic numerical stability and consistency principles.
The Feature Agent extracts numerical features from both the problem and the proposed schemes, and the Selector Agent scores and ranks these candidates, further filtering out ill-designed or nonphysical plans before selecting the top-$k$ for execution.

\subsection{Coarse-to-Fine Execution}

Debugging LLM-generated code directly on high-resolution grids is computationally wasteful.
We decouple logic debugging from stability validation through a coarse-to-fine strategy.
In the coarse-grid phase, the solver runs at reduced resolution, and the Critic Agent fixes logic issues (syntax errors, shape mismatches).
Once logic validation passes, the code is promoted to the high-resolution grid, where failures are treated as numerical stability issues and addressed by adjusting the time step.

If repair attempts exceed the retry limit $M$ at either stage, the system triggers a Fresh Restart.
The current code is discarded and the Coder Agent generates a new implementation from scratch, enabling the system to escape failed code paths.
For large-scale temporal simulations, the Coder Agent is instructed to store solution snapshots only at sparse intervals to avoid memory exhaustion.

\subsection{Verification and Analysis}

Verifying solver correctness is a core challenge in automated PDE solving.
Let $u$ denote the numerical solution, $u^*$ the analytic solution (when available), and $\mathcal{L}$ the PDE operator.
When an explicit analytic solution exists, we compute the relative $L_2$ error;
when no analytic solution is available, we evaluate the relative PDE residual;
and for implicit analytic relations (e.g., conservation laws $F(u)=0$), we
measure the relative implicit residual. These three errors are defined respectively as
\begin{equation}\label{eq: metrics}
e_{L_2} = \frac{\|u - u^*\|_{L^2(\Omega)}}{\|u^*\|_{L^2(\Omega)}+\epsilon}, \quad
e_{\mathrm{res}} = \frac{\|\mathcal{L}(u) - f\|_{L^2(\Omega)}}{\|f\|_{L^2(\Omega)} + \epsilon}, \quad
e_{\mathrm{impl}} = \frac{\|F(u)\|_{L^2(\Omega)}}{\|F_{\mathrm{ref}}\|_{L^2(\Omega)} + \epsilon}, \text{where }\epsilon = 10^{-12}
\end{equation}
Generated solvers are required to compute and return residuals, and the system enforces validity checks on these values.
Finally, a Reasoning Agent generates theoretical analysis for the best-performing scheme.

\section{Experiments \& Results}

\subsection{Experimental Setup}
\textbf{Benchmark:}
We evaluate our framework on two benchmarks:
\textbf{(1) CodePDE Benchmark.} To enable fair comparison with existing neural network solvers and LLM-based methods, we adopt the benchmark proposed by CodePDE, which comprises 5 representative PDEs: 1D Advection, 1D Burgers, 2D Reaction-Diffusion, 2D Compressible Navier-Stokes (CNS), and 2D Darcy Flow. These problems span linear and nonlinear equations, elliptic and time-dependent types, as well as diverse boundary conditions and levels of numerical stiffness.
\textbf{(2) Our Benchmark.} To more comprehensively assess the generality of our framework, we construct a large-scale benchmark suite containing 200 different PDEs, covering a wide range of common PDE families (Advection, Burgers, Fokker-Planck, Heat, Maxwell, Poisson, etc.). The PDEs in our benchmark range from 1D to 5D in spatial dimension and span elliptic, parabolic, hyperbolic types as well as PDE systems. They include linear and nonlinear, stiff and non-stiff, steady-state and time-dependent problems, with Dirichlet, Neumann, and periodic boundary conditions.

\textbf{Numerical Settings:}
The Planner Agent generates 10 candidate solver schemes and scores each one for every PDE problem.
The top-5 schemes are passed to the Coder Agent for implementation.
We set the maximum number of retries for code generation, coarse-grid execution, and high-resolution execution to 2, 4, and 6, respectively.
The maximum wall-clock time for each coarse-grid or high-resolution run is 120 seconds.

\textbf{Evaluation Metrics:}
We evaluate solver accuracy using the three metrics defined in Section 2.3 ($e_{L_2}$, $e_{\text{impl}}$, $e_{\text{res}}$)~\ref{eq: metrics}, depending on the available reference information.
We also report execution time, defined as the wall-clock time from solver generation to the first successful evaluation.

\subsection{Results and Analysis}
\begin{table}[!ht]
  \centering
  \caption{\small \textbf{nRMSE (normalized root mean square error) comparison with neural network baselines and CodePDE.} All LLM-based methods (CodePDE and Ours) use GPT-4.1. CodePDE results are obtained under the Reasoning + Debugging + Refinement setting (best of 12).}
  \label{tab:comparison_codepde}
  \resizebox{\linewidth}{!}{%
  \begin{tabular}{c|ccccc|c}
  \toprule
  \midrule
  nRMSE ($\downarrow$) & Advection & Burgers & React-Diff & CNS & Darcy & Geom.\ Mean \\
  \midrule
  U-Net & \cerr{5.00e-2} & \cerr{2.20e-1} & \cerr{6.00e-3} & \cerr{3.60e-1} & $-$ & $-$ \\
  FNO & \cerr{7.70e-3} & \cerr{7.80e-3} & \cerr{1.40e-3} & \cerr{9.50e-2} & \cerr{9.80e-3} & \cerr{9.52e-3} \\
  PINN & \cerr{7.80e-3} & \cerr{8.50e-1} & \cerr{8.00e-2} & $-$ & $-$ & $-$ \\
  ORCA & \cerr{9.80e-3} & \cerr{1.20e-2} & \cerr{3.00e-3} & \cerr{6.20e-2} & $-$ & $-$ \\
  PDEformer & \cerr{4.30e-3} & \cerr{1.46e-2} & $-$ & $-$ & $-$ & $-$ \\
  UPS & \cerr{2.20e-3} & \cerr{3.73e-2} & \cerr{5.57e-2} & \cerr{4.50e-3} & $-$ & $-$ \\
  \midrule
  CodePDE & \cerr{1.01e-3} & \cerr{3.15e-4} & \cerr{1.44e-1} & \cerr{1.53e-2} & \cerr{4.88e-3} & \cerr{5.08e-3} \\
  \midrule
  Central Difference (Ill-designed)
 & \cerr{7.05e12} & \cerr{1.64e-2} & \cerr{1.23e-1} & \cerr{3.85} & \cerr{2.34e-1} & $-$ \\
  Ours & \cerrb{4.18e-14} & \cerrb{1.79e-5} & \cerrb{8.98e-7} & \cerrb{1.82e-4} & \cerrb{4.84e-13} & \cerrb{9.00e-9} \\
  \midrule
  \bottomrule
  \end{tabular}
  }
\end{table}

We select 24 representative problems from our 200-PDE benchmark suite, spanning 1D to 5D and covering elliptic, parabolic, and hyperbolic types (full results in Appendix Table~\ref{tab:split_results_cont}).
Among the 19 problems with explicit analytic solutions, 11 achieve relative $L_2$ errors of $10^{-6}$ or better, with Poisson (\cerr{5.41e-16}) and Helmholtz 2D (\cerr{3.50e-16}) reaching near machine precision.
Biharmonic (\cerr{6.14e-1}) and 5D Helmholtz (\cerr{9.8e-1}) are notable failure cases, indicating limited capability on fourth-order and high-dimensional PDEs.
End-to-end runtimes fall between 20 and 130 seconds for most problems.
A step-by-step walkthrough of the full pipeline on one example problem is provided in Appendix~\ref{sec:example-output}.

Table~\ref{tab:comparison_codepde} compares our method with six neural network baselines, CodePDE, and an ill-designed solver on the five CodePDE benchmark problems; all baseline results are reproduced from~\citet{li2025codepde}.
Our method achieves the lowest nRMSE on all five problems, with a geometric mean of \cerr{9.00e-9}, approximately six orders of magnitude below CodePDE (\cerr{5.08e-3}) and the Fourier Neural Operator (FNO, \cerr{9.52e-3}).
As a reference point, this ill-designed central finite-difference baseline,
obtained from an existing online implementation and applied naively without
stability safeguards, yields extremely large nRMSE across the five PDEs,
reaching \cerr{7.05e12} on the advection case. This counterexample highlights
the importance of stability-aware plan generation and selection in our
pipeline for preventing such ill-designed solvers from being executed.
Analysis of the selected schemes across all 24 problems (see Appendix Table~\ref{tab:scheme_selection}) reveals a consistent pattern: the Planner Agent selects Fourier spectral methods for periodic-boundary problems, finite difference or finite element methods for Dirichlet-boundary parabolic problems, and Chebyshev spectral methods for Dirichlet-boundary elliptic problems.

\section{Conclusion}
The Planner and Selector agents embed stability- and
consistency-aware numerical reasoning into the generation process, enabling
the pipeline to detect and exclude ill-designed or nonphysical solver configurations prior to execution. Through a subsequent coarse-to-fine
execution strategy and residual-based self-verification, the system then
performs end-to-end solver construction and quality assessment without requiring analytical solutions.
Experiments on 24 benchmark PDEs indicate that the framework selects numerical schemes consistent with PDE structural properties (e.g., spectral methods for periodic domains, finite differences for Dirichlet boundaries), and achieves lower error than both neural network baselines and CodePDE on the majority of the CodePDE benchmark problems.
The framework still exhibits limited accuracy on high-dimensional ($\geq$5D) and high-order PDEs, and our evaluation covers only regular domains.
The system is also coupled to a single LLM (GPT-4.1), and the generated code lacks formal convergence or stability guarantees.

\subsubsection*{Acknowledgments}
The authors were partially supported by the US National Science Foundation under awards IIS-2520978, GEO/RISE-5239902, the Office of Naval Research Award N00014-
23-1-2007, DOE (ASCR) Award DE-SC0026052, and the DARPA D24AP00325-00. Approved for public release; distribution is unlimited.


\bibliography{iclr2026_conference}
\bibliographystyle{iclr2026_conference}

\newpage
\appendix
\section{Related Work}
\label{sec:related-work}

\paragraph{Classical Numerical Methods.}
Classical numerical analysis remains the foundation for solving PDEs.
The finite difference method approximates derivatives using grid-based differences~\citep{leveque2007finite}.
The finite element method represents solutions over mesh elements~\citep{zienkiewicz2013finite}.
Spectral methods expand solutions in global basis functions~\citep{canuto2007spectral}.
Despite their mathematical rigor, constructing effective solvers typically requires substantial expertise in discretization design and stability verification, motivating interest in automated solver construction.

\paragraph{Neural and Data-Driven PDE Solvers.}
Scientific machine learning has introduced neural-network-based approaches for approximating PDE solutions, including PINNs~\citep{raissi2019physics} and neural operators~\citep{lu2021learning,li2020fourier}.
Subsequent work explores Transformers~\citep{cao2021choose}, message-passing neural networks~\citep{brandstetter2022message}, state-space models~\citep{zheng2024alias,buitrago2025on}, and pretrained multiphysics foundation models~\citep{shen2024ups,subramanian2023towards,mccabe2024multiple}.

\paragraph{LLMs for Scientific Computing and PDE Automation.}
Large language models have demonstrated strong capability in generating executable scientific code across chemistry~\citep{bran2023chemcrow}, physics~\citep{arlt2024metadesigningquantumexperimentslanguage}, mathematics~\citep{wang2023mathcoderseamlesscodeintegration}, and computational biology~\citep{tang2024biocoderbenchmarkbioinformaticscode}.
Agentic reasoning frameworks extend these capabilities through planning and structured tool interaction~\citep{romera2024mathematical,Ma2024LLMAS,jiang2025aide,zhou2025autonomous}.
FunSearch~\citep{romera2024mathematical} demonstrates program search for mathematical structure discovery, while PDE-Controller~\citep{soroco2025pde} explores LLM-driven autoformalization for PDE control.
Closer to automated PDE solving, neural solver design frameworks construct PINNs via multi-agent reasoning~\citep{he2025lang,jiang2025agenticsciml}, tool-oriented systems orchestrate libraries such as FEniCS~\citep{liu2025pde,wu2025automated}, and code-generation paradigms synthesize candidate solvers~\citep{li2025codepde}.

\newpage
\section{Full Benchmark Results}
\label{sec:full-results}
Table~\ref{tab:split_results_cont} reports per-problem accuracy and runtime for all 24 benchmark PDEs.

\begin{table}[!ht]
\caption{Evaluation of proposed framework across 24 benchmark PDEs. The upper block reports relative $L_2$ error for problems with known analytic solutions; the lower block reports relative residual error.}
\label{tab:split_results_cont}
\centering
\footnotesize
\setlength{\tabcolsep}{4pt}
\begin{tabular}{cccc}
\toprule
\textbf{PDE} &
\textbf{Dim} &
\textbf{Error} &
\textbf{Runtime (s)} \\
\midrule
\multicolumn{4}{l}{\textit{Explicit analytic solution available \quad (Relative $L_2$ error)}} \\
\midrule
Advection            & 2 & \cerr{1.13e-13} & 29.8  \\
Allen-Cahn           & 1 & \cerr{2.23e-4}  & 19.8  \\
Biharmonic           & 2 & \cerr{6.14e-1}  & 89.3  \\
Convection Diffusion & 2 & \cerr{8.57e-3}  & 34.6  \\
Euler                & 1 & \cerr{5.21e-14} & 26.0  \\
Heat                 & 1 & \cerr{3.21e-7}  & 97.4  \\
Heat                 & 2 & \cerr{1.50e-4}  & 228.1 \\
Helmholtz            & 2 & \cerr{3.50e-16} & 66.3  \\
Helmholtz            & 5 & \cerr{9.8e-1}   & 65.8  \\
KdV                  & 1 & \cerr{2.36e-7}  & 52.2  \\
Laplace              & 2 & \cerr{1.24e-5}  & 85.9  \\
Maxwell              & 3 & \cerr{1.00e-3}  & 126.1 \\
Navier--Stokes       & 2 & \cerr{8.08e-6}  & 64.5 \\
Poisson              & 2 & \cerr{5.41e-16} & 68.9  \\
Reaction Diffusion   & 2 & \cerr{9.88e-6}  & 199.5 \\
Schr\"{o}dinger      & 1 & \cerr{5.40e-14} & 32.2  \\
Shallow Water        & 1 & \cerr{1.67e-10} & 18.5  \\
Vorticity            & 2 & \cerr{3.32e-4}  & 54.1  \\
Wave                 & 1 & \cerr{8.34e-10} & 73.1  \\
\midrule
\multicolumn{4}{l}{\textit{Implicit analytic solution available \quad (Relative implicit residual error)}} \\
\midrule
Burgers (inviscid)   & 1 & \cerr{5.65e-4}  & 23.4  \\
\midrule
\multicolumn{4}{l}{\textit{No analytic solution \quad (Relative residual error)}} \\
\midrule
Burgers (viscous)    & 1 & \cerr{8.95e-14} & 63.1  \\
Cahn--Hilliard       & 1 & \cerr{9.88e-4}  & 114.9 \\
Fokker--Planck       & 2 & \cerr{2.24e-3}  & 44.3  \\
Gray--Scott          & 2 & \cerr{1.10e-3}  & 23.7  \\
\bottomrule
\end{tabular}
\end{table}

\newpage
\section{Pipeline Walkthrough: 2D Advection}
\label{sec:example-output}
We walk through the full pipeline output for 2D Advection ($u_t + c_x u_x + c_y u_y = 0$, periodic BCs, $c_x{=}0.3$, $c_y{=}0.2$).

\paragraph{Step 1: Planner Agent.}
The Planner generates 10 candidate schemes spanning spectral, finite difference (FD), finite volume (FV), and finite element (FEM) methods with various time integrators (RK4: classical fourth-order Runge-Kutta; IMEX: implicit-explicit; ETDRK4: exponential time differencing RK4).
The Selector Agent scores each based on expected accuracy, stability, and cost.
Table~\ref{tab:plans_scored} lists all candidates.

\begin{table}[!ht]
\caption{Candidate schemes generated by the Planner Agent and scored by the Selector Agent for the 2D Advection problem.}
\label{tab:plans_scored}
\centering
\small
\begin{tabular}{lrcl}
\toprule
\textbf{Plan} & \textbf{Score} & \textbf{Method} & \textbf{Rationale (summary)} \\
\midrule
Spectral (RK4, high-res)     & 90 & Spectral Fourier & Optimal for smooth periodic advection \\
FD (WENO3+RK3, high-res)     & 85 & Finite Difference & High-order upwind, conservative \\
Spectral (ETDRK4, med-res)   & 80 & Spectral Fourier & Good accuracy-cost balance \\
FV (MUSCL+RK2, med-res)      & 75 & Finite Volume    & Limiter controls oscillations \\
FD (semi-Lagrangian, med-res) & 70 & Finite Difference & Large time steps, second-order \\
FEM (IMEX, med-res)           & 60 & Finite Element   & Stable but diffusion treatment irrelevant \\
FD (upwind, low-res)          & 55 & Finite Difference & Stable but first-order accuracy \\
FD (Crank-Nicolson, med-res)  & 50 & Finite Difference & No upwinding, oscillation risk \\
FV (upwind, low-res)          & 50 & Finite Volume    & Stable but low accuracy \\
FEM (backward Euler, med-res) & 45 & Finite Element   & Implicit cost unjustified \\
\bottomrule
\end{tabular}
\end{table}

\paragraph{Step 2: Coder + Critic Agents.}
The top-5 plans are implemented and executed through the coarse-to-fine pipeline.
Table~\ref{tab:plans_evaluated} reports the execution results.

\begin{table}[!ht]
\caption{Execution results for the top-5 candidate schemes on the 2D Advection problem.}
\label{tab:plans_evaluated}
\centering
\small
\begin{tabular}{lcccc}
\toprule
\textbf{Plan} & \textbf{Residual $L_2$} & \textbf{Runtime (s)} & \textbf{Attempts} & \textbf{Restarts} \\
\midrule
Spectral (RK4, high-res)      & \cerr{1.75e-3}  & 23.8 & 2 & 0 \\
FD (WENO3+RK3, high-res)      & \cerr{3.18e4}   & 57.5 & 4 & 0 \\
Spectral (ETDRK4, med-res)    & \cerr{8.02e-15} & 35.3 & 4 & 0 \\
FV (MUSCL+RK2, med-res)       & \cerr{1.94e-1}  & 33.2 & 2 & 0 \\
FD (semi-Lagrangian, med-res) & \cerr{2.27e-2}  & 15.8 & 2 & 0 \\
\bottomrule
\end{tabular}
\end{table}

\paragraph{Step 3: Final Selection.}
The Selector Agent chooses \textbf{Spectral (ETDRK4, med-res)} based on its residual of \cerr{8.02e-15} (near machine precision) at a moderate runtime of 35.3\,s.
The high-resolution spectral plan, despite scoring highest in planning, produces a larger residual (\cerr{1.75e-3}), likely due to time-stepping error at coarser $\Delta t$.
The FD plan diverges entirely (residual \cerr{3.18e4}).
This example illustrates how the pipeline's evaluate-then-select strategy can override initial scoring when execution results differ from expectations.

\newpage
\section{Scheme Selection Results}
\label{sec:scheme-selection}

Table~\ref{tab:scheme_selection} lists the numerical scheme automatically selected by the Planner Agent for each benchmark PDE.
The schemes are grouped by boundary condition type.
For periodic-boundary problems, the pipeline consistently selects Fourier spectral methods.
For Dirichlet-boundary parabolic problems, finite difference (FD) or finite element methods (FEM) with implicit time stepping are preferred.
For Dirichlet-boundary elliptic problems, Chebyshev spectral methods are selected.

\begin{table}[!ht]
\caption{Numerical schemes selected by the Planner Agent for each benchmark PDE.}
\label{tab:scheme_selection}
\centering
\small
\begin{tabular}{lccll}
\toprule
\textbf{PDE} & \textbf{Dim.} & \textbf{BC} & \textbf{PDE Type} & \textbf{Selected Scheme} \\
\midrule
\multicolumn{5}{l}{\textit{Periodic boundary conditions}} \\
\midrule
Advection            & 2 & Periodic  & Hyperbolic & Spectral Fourier (RK4) \\
Convection Diffusion & 2 & Periodic  & Parabolic  & Spectral Fourier (IMEX) \\
Schr\"{o}dinger      & 1 & Periodic  & Dispersive & Spectral Fourier (Split-Step) \\
Navier--Stokes       & 2 & Periodic  & Parabolic  & FEM (IMEX) \\
Shallow Water        & 1 & Periodic  & Hyperbolic & FD (explicit) \\
\midrule
\multicolumn{5}{l}{\textit{Dirichlet boundary conditions, parabolic}} \\
\midrule
Allen-Cahn           & 1 & Dirichlet & Parabolic  & FD (Crank-Nicolson) \\
Burgers (viscous)    & 1 & Dirichlet & Parabolic  & FD (implicit) \\
Heat                 & 1 & Dirichlet & Parabolic  & FEM (Crank-Nicolson) \\
Heat                 & 2 & Dirichlet & Parabolic  & FD (Crank-Nicolson) \\
Reaction Diffusion   & 2 & Dirichlet & Parabolic  & FD (IMEX) \\
\midrule
\multicolumn{5}{l}{\textit{Dirichlet boundary conditions, elliptic}} \\
\midrule
Helmholtz            & 2 & Dirichlet & Elliptic   & Spectral Chebyshev \\
Laplace              & 2 & Dirichlet & Elliptic   & Spectral Chebyshev \\
Poisson              & 2 & Dirichlet & Elliptic   & Spectral Chebyshev \\
\midrule
\multicolumn{5}{l}{\textit{Dirichlet boundary conditions, hyperbolic}} \\
\midrule
Wave                 & 1 & Dirichlet & Hyperbolic & Spectral (explicit) \\
\bottomrule
\end{tabular}
\end{table}

\subsubsection*{AI Usage}
This work used large language models for language polishing,
formatting assistance, and limited code suggestions.

\end{document}